\documentclass[runningheads]{llncs}
\usepackage[T1]{fontenc}
\usepackage{CJKutf8}
\usepackage{graphicx}

\usepackage{epsfig}
\usepackage{booktabs}
\usepackage{algorithmic}
\usepackage{textcomp}

\usepackage{framed,multirow}
\usepackage{xcolor}
\usepackage{cite}
\usepackage{url}

\newcommand{\ja}[1]{%
  \begin{CJK}{UTF8}{ipxm}#1\end{CJK}%
}
\begin{document}
\title{
JaPOC: Japanese Post-OCR Correction Benchmark using Vouchers
}
%
%
\author{Masato Fujitake\inst{1}\orcidID{0000-0001-7702-499X}
}
\authorrunning{M. Fujitake}
%
\institute{
FA Research, 
Fast Accounting Co., Ltd.
, Japan\\
\email{fujitake@fastaccounting.co.jp}\\
}
\maketitle              
\begin{abstract}
In this paper, we create benchmarks and assess the effectiveness of error correction methods for Japanese vouchers in OCR (Optical Character Recognition) systems.
It is essential for automation processing to correctly recognize scanned voucher text, such as the company name on invoices.
However, perfect recognition is complex due to the noise, such as stamps. 
Therefore, it is crucial to correctly rectify erroneous OCR results.
However, no publicly available OCR error correction benchmarks for Japanese exist, and methods have not been adequately researched.
In this study, we measured text recognition accuracy by existing services on Japanese vouchers and developed a post-OCR correction benchmark. 
Then, we proposed simple baselines for error correction using language models and verified whether the proposed method could effectively correct these errors. 
In the experiments, the proposed error correction algorithm significantly improved overall recognition accuracy.

\end{abstract}
\section{Introduction}
\begin{figure}[htp]
	\centering
	\includegraphics[width=1.00\columnwidth, keepaspectratio]{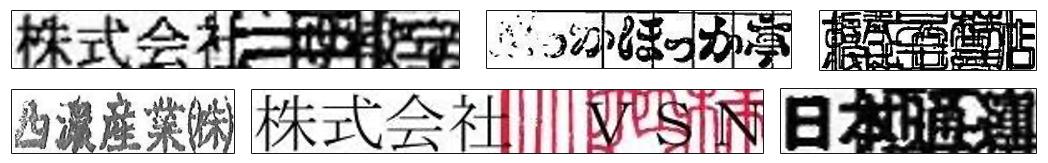}
	\caption{
   Sample examples of Japanese text recognition images in vouchers. 
 Due to the convention of stamping seals on vouchers, text images of vouchers tend to be difficult for OCR to read. 
	}
	\label{fig:dataset_samples}
\end{figure}

In business automation scenarios, a system that can accurately extract text from corporate document images, such as invoices, is needed. 
Despite the growth in digital documents, paper-based vouchers still exist. 
Thus, OCR (Optical Character Recognition) technology~\cite{zhang2022testr, fujitake2023diffusionstr, fujitake2023dtrocr} is essential to read text from scanned vouchers. 
Various OCR services~\cite{googlecloudvision, fa_robota} and research~\cite{zhang2022testr, fujitake2023a3s, fujitake2024jstr} for higher accuracy have been conducted in recent years due to the necessity of this technology. 
However, OCR is affected by various fonts and image conditions and may not be able to read the text accurately. 
In addition, in Japanese accounting routines, it is customary to stamp seals around company names on the documents to increase processing reliability against forgery. 
As a result, noisy text images are sometimes created, as shown in figure~\ref{fig:dataset_samples}, and the recognition accuracy tends to be significantly degraded by the seal impression. 
Therefore, correcting the errors after OCR results is necessary to extract accurate information.

Post-OCR correction in Japanese OCR has been studied in old books and ancient documents. 
The previous studies consisted of two components: detecting error locations and selecting correction characters for error candidates. 
In the detection, each character is judged whether it is an error or not based on trigrams. 
If the probability value in the language models is smaller than a threshold value, the character is considered as an error candidate. 
Then, when correcting the error candidates, a character string with a high probability of occurrence is substituted for each character position in the error correction candidates ~\cite{takeuchi1999postocr}. 
For improved accuracy, a method has been proposed to utilize not only local information but also the global context of the entire OCR text~\cite{masuda2015contextualjapostocr}.

Prior work mainly focused on transcribing books and old documents, and benchmarks for corporate documents such as vouchers have yet to be built openly available.
In addition, the previous studies are complex and manual in terms of setting thresholds and so on. 

To address these issues, we construct a publicly available benchmark for post-OCR correction in Japanese vouchers.
They contain various information such as company name, amount, date, etc., each subject to OCR error. 
However, information such as amounts and dates do not have linguistic meanings as a single piece of information, and effective error correction is considered to be difficult. 
Therefore, in this study, we focus on the company name as a supplier name and construct a benchmark with real-world document images and multiple OCR services. 
Then, using the benchmark, we constructed baselines using language models and a rule-based method, and evaluated the baselines. 
Experiments showed that post-OCR correction models fine-tuned from appropriately chosen pre-trained language models provide significant accuracy improvements over rule-based models.
Our contributions are as follows:

\begin{itemize}
    \item We proposed JaPOC, a post-OCR correction benchmark based on Japanese vouchers. 
It contains real-world errors based on two different OCR services, which provide various characteristics and would help as a basis for future research. 
 The benchmarks are available.\footnote{
    \url{
https://github.com/FastAccounting/ocr_correction_benchmark
    }}.
    
    \item We provided baselines for the benchmark using 
 language models and a rule-based analysis method. 
\end{itemize}

\section{Related Works} \label{sec:relatedwork}
\subsection{Post-OCR Correction in Japanese}
Japanese OCR error correction tasks have been studied for books based on a two-step method: detection and selection~\cite{takeuchi1999postocr}, and an improved method for selecting correction candidates using global information has been proposed ~\cite{masuda2015contextualjapostocr }. 
A misrecognized OCR detection method using the language model BERT is also proposed ~\cite{sha2023japanese_ocr}. 
Unlike previous works, this study proposes an approach that directly corrects text with a single language model and creates a public-available benchmark for future work.

\subsection{Pretrained Model: T5}
Recent studies have focused on pre-training models trained on large amounts of unlabeled data and performing well on many tasks. 
One of these models, T5~\cite{raffel2020t5}, views various tasks in natural language processing as a sequence-to-sequence transformation process and improves several task performance by utilizing Transformer~\cite{vaswani2017transformer}. 
In this study, we define post-OCR correction as a sequence-to-sequence transformation and apply the model for this task.

\section{Benchmark Construction} \label{sec:benchmark}
\begin{table}[tb]
\centering
\caption{
OCR recognition accuracy by various models and services
}
\label{tab:ocr_accuracy}
\resizebox{0.95\columnwidth}{!}{%
\small
\begin{tabular}{c|c|c }
\toprule
Method      & Acc w/o post-process  & Acc w/ post-process \\ \hline
Japanese OCR~\cite{detomo2021japaneseocr} & 24.6 & 26.6 \\
Google Vision~\cite{googlecloudvision} & 42.6 & 72.0 \\
Fast Accounting Robota~\cite{fa_robota} & 97.0 & 99.1 \\
\bottomrule
\end{tabular}
}
\end{table}

\begin{table*}[tb]
\centering
\caption{
Example of recognition result. 
}
\label{tab:ocr_result}
\resizebox{1.00\columnwidth}{!}{%
\small
\begin{tabular}{c|c|c |c}
\toprule
Ground Truth      &  Japanese OCR & Vision API & Robota API \\ \hline
\ja{向島運送株式会社}      &  \ja{河島選選が出てきます.}  & \ja{向島運送株会社} & \ja{向島運送株式会社} \\ \hline 
\ja{株式会社DAISHIZEN}      & \ja{株式会社DAISHIZEN}  & \ja{\#DAISHIZEN} & \ja{株式会社DAISHIZEN} \\ \hline 
\ja{日本ロジテム株式会社}&  \ja{日本の日が流れましたが,}  & \ja{日本株式会社} & \ja{日本ロジテム株式会社} \\ \hline 
\ja{株式会社横浜ファーマシー} & \ja{株式会社機浜ファーナー}  & \ja{株式会社横浜ファーマジ} & \ja{株式会社横浜ファーマント}  \\  
\bottomrule
\end{tabular}
}
\end{table*}

In this section, we first check the performance of current OCR models and services on real-world data to create a post-OCR correction benchmark and then use the results to create a benchmark.

\subsection{OCR Evaluation}

We have constructed an OCR error correction dataset that includes the effects of seals on Japanese vouchers. 
We prepared a set of images by cropping the textual regions of company name items using Japanese invoices with permission, as shown in Fig.~\ref{fig:dataset_samples}. 
We randomly sampled 11,000 voucher images. 
Then, we created pairs of annotated data and the OCR model's text recognition results for the images. 
Specifically, all images were annotated with Ground Truth (GT) by a human specializing in annotating text recognition images. 
For OCR, we used three models and services to evaluate the current performance of text recognition. 
The first is ``Japanese OCR''~\cite{detomo2021japaneseocr}, which is an openly available Transformer-based Japanese text recognition model. 
The second is Vision API~\cite{googlecloudvision} provided by Google, which is a commercial general-purpose text recognition. 
The third model is Robota API~\cite{fa_robota}, a commercial accounting-specific OCR provided by Fast Accounting. 
The OCR accuracy of each model was evaluated in terms of word-level accuracy. 
That is, if all the characters in GT and the OCR result are matched, the result is True. Otherwise, it is False. 
Alphabetic, alphanumeric, and space characters were treated as half-width characters. 
Accuracy of each method is shown in Table ~\ref{tab:ocr_accuracy}. 
Recognition of whitespace, line feed, and character code differs depending on the model and service. 
Although whitespaces and line feeds may have essential meanings, completing them only from textual information is challenging.
Considering OCR error correction with only text, the accuracy is calculated separately after standardization post-processes such as removing them. 
After the post-processing was applied, the OCR accuracies of Robota API, Vision API, and Japanese OCR were 99.1, 72.0, and 26.6 percent, respectively.

Table~\ref{tab:ocr_result} shows GT and its OCR results for the random samples. 
We confirm that ``Japanese OCR'' tends to return recognition results similar to those of Japanese sentences, and that it is affected by noise that prevents correct recognition. 
In addition, the Vision API recognizes legible characters correctly, but some characters are missing. 
Robota API has fewer missing characters. 

\subsection{Development of Post-OCR Correction Benchmark}
Based on the previous results, we have constructed a post-OCR correction benchmark. 
The OCR error correction task in this work is to take a text sequence that may contain errors as input and output the corrected one as output. 
If there are no errors as input, the correct text is output as it is. 
Based on the above task setup, we constructed benchmarks using the text recognition results and GT. 
The recognition results of each service are used as input, and the output is estimated as GT. 
We removed a person's name, such as a company representative, from the dataset from the perspective of personal information protection.

Since the accuracy of ``Japanese OCR'' is low and it is considered difficult to qualitatively estimate the correct character from the output results, we used the results of Vision API and Robota API as each set of benchmark. 
Therefore, the benchmark consists of two sets. 
We apply the same evaluation metrics for post-OCR correction as for OCR recognition accuracy, and measure each set's score and the average.
The training, evaluation, and test sets are randomly separated based on the images. 
Table~\ref{tab:dataset_distribution} shows each set's number of True and False samples for train, validation, and test. 
True means that the GT and OCR results match, and false means that the samples contain errors. 
Although the Robota and Vision sets share the same images, the proportion of True and False samples is different because the accuracy of their recognition results is different. 
Thus, we constructed a Japanese post-OCR correction benchmark using these two sets with different percentages and error tendencies.
\begin{table}[tb]
\centering
\caption{
Number of data samples for training, evaluation, and test sets.
}
\label{tab:dataset_distribution}
\resizebox{0.9\columnwidth}{!}{%
\small
\begin{tabular}{c|r|r |r |r }
\toprule
Set      &  Vision True  & Vision False & Robota True & Robota False\\ \hline
Train    &    6,817    &  2,640   & 9,371 & 86 \\ \hline
val    &    353    &  145   & 490 & 8 \\ \hline
Test    &      717  &  279   & 983 & 13\\ 
\bottomrule
\end{tabular}
}
\end{table}

\section{Method} \label{sec:method}
\subsection{T5 Based Approach}
We evaluate the performance of language models in correcting OCR errors. 
We conducted experiments using T5 as a suitable model for evaluating post-OCR correction. 
T5 is an encoder-decoder language model based on Transformer, and its structure allows for sequence-to-sequence conversion. 
Therefore, it is possible to take input strings that may contain errors, which is the task setting of this study, and produce corrected output strings.
We used two Japanese pre-trained weights for T5.
Then, we fine-tuned and trained the pre-trained models on benchmarks. 

One is the pre-trained Japanese T5 published by Megagon Labs, with a vocabulary size of 32k~\cite{megagon2021t5ja}. 
In this study, we denote it as T5(~\textsubscript{Megagon}). 
The other model is the pre-trained Japanese model T5~\cite{retrieva2021t5ja} published by Retrieva Inc. 
It is based on T5X, a slightly modified version of T5.
In this study, we denote the model as T5\textsubscript{Retrieva}. 
In all cases, mC4 and Wikipedia are used as their pre-training. 

Fine-tuning them to the post-OCR correction task was performed in the framework of a sequence transformation, using the OCR error correction candidate text as input to the T5 encoder and the decoder output as error-corrected text. 
Thus, the model is trained to output the corrected result if the input has errors and to output the input as it is if there are no errors.

We basically followed the training script \footnote{\url{https://github.com/huggingface/transformers/blob/main/examples/pytorch/summarization/run_summarization. py}}. 
The same hyperparameters were used for both models with a batch size of 32, a maximum input-output length of 64, a learning rate of 5e-5, an iteration of 15,000, and a gradient accumulation of 2 steps. 
The quantitative evaluation is the average of three training runs. 
In the preliminaries, we confirmed that for the pre-trained models, the results without fine-tuning yielded a score of 0 in both cases.

\subsection{Rule Based Approach}
To compare the performance of the language model approach, we also constructed a rule-based model. 
This approach uses the National Tax Agency's corporate database to perform a search by edit distance. 
If a candidate text can be found using the distance, it is modified and output. 
Specifically, we created a corporate database based on the database\footnote{\url{https://www.houjin-bangou.nta.go.jp/}}, performed preprocessing and deduplication, and produced a list of about 3.7 million corporations. 
The preprocessing is based on the standardization process used when creating the benchmarks. 
The Levenshtein distance is used as the edit distance. 
The strings with the smallest distance are retrieved from the database as candidates, and scores are calculated for comparison. 
The score consists of two parts, one is the edit distance itself, and the other is the ratio of the edit distance normalized to the string. 
The ratio is calculated by dividing the edit distance by the maximum length of the computed text. 
If the edit distance is less than two and the ratio is less than 0.30, the candidate text is output as the corrected text; otherwise, the input text is output. 

\section{Experiments} \label{sec:experiments}
\begin{table*}[tb]
\centering
\caption{
Results after OCR correction by the language model on the test set.
}
\label{tab:correction_result}
\resizebox{1\columnwidth}{!}{%
\small
\begin{tabular}{c|c|c|c |c}
\toprule
Test Set & Method &  Ground Truth      &  Before correction  & After correction \\ \hline
Vision & T5\textsubscript{Retrieva} & \ja{株式会社DAISHIZEN} &  \ja{\#DAISHIZEN} & \ja{株式会社DAISHIZEN} \\
Vision & T5\textsubscript{Retrieva} & \ja{東洋エアゾール工業株式会社} &  \ja{東洋工業株式会社} & \ja{東洋エアゾール工業株式会社} \\
Vision & T5\textsubscript{Retrieva} & \ja{株式会社優} & \ja{株式会社三菱} & \ja{株式会社三菱ふそう}\\
Vision & T5\textsubscript{Retrieva} & \ja{ユニックス} & \ja{Z} & \ja{ほっかほっか亭}\\
Vision & T5\textsubscript{Retrieva} & \ja{株式会社クリーン・アシスト} & \ja{株式会社クリーン・アンプト} & \ja{株式会社クリーン・アップト}\\

Vision & T5\textsubscript{Megagon} & \ja{株式会社DAISHIZEN} &  \ja{\#DAISHIZEN} & \ja{DAISHIZEN} \\
Vision & T5\textsubscript{Megagon} & \ja{東洋エアゾール工業株式会社} & \ja{東洋工業株式会社} & \ja{東亜工業株式会社}\\
Vision & T5\textsubscript{Megagon} & \ja{株式会社優} & \ja{株式会社三菱} & \ja{三菱}\\
Vision & T5\textsubscript{Megagon} & \ja{ユニックス} & \ja{Z} & \ja{会社}\\
Vision & T5\textsubscript{Megagon} & \ja{株式会社クリーン・アシスト} & \ja{株式会社クリーン・アンプト} & \ja{クリーン・アンプト}\\

Robota & T5\textsubscript{Retrieva} & \ja{株式会社DAISHIZEN} &  \ja{株式会社DAISHIZEN} & \ja{株式会社DAISHIZEN} \\
Robota & T5\textsubscript{Retrieva} & \ja{東洋エアゾール工業株式会社} & \ja{東洋エアゾール工業株式会社} & \ja{東洋エアゾール工業株式会社} \\
Robota & T5\textsubscript{Retrieva} & \ja{株式会社優} & \ja{株式会社優} & \ja{株式会社優}\\
Robota & T5\textsubscript{Retrieva} & \ja{ユニックス} & \ja{ユニックス} & \ja{ユニックス}\\
Robota & T5\textsubscript{Retrieva} & \ja{株式会社クリーン・アシスト} & \ja{株式会社クリーン・アジスト} & \ja{株式会社クリーン・アシスト}\\

Robota & T5\textsubscript{Megagon} & \ja{株式会社DAISHIZEN} &  \ja{株式会社DAISHIZEN} & \ja{DAISHIZEN} \\
Robota & T5\textsubscript{Megagon} & \ja{東洋エアゾール工業株式会社} & \ja{東洋エアゾール工業株式会社} & \ja{エアゾール工業株式会社} \\
Robota & T5\textsubscript{Megagon} & \ja{株式会社優} & \ja{株式会社優} & \ja{優}\\
Robota & T5\textsubscript{Megagon} & \ja{ユニックス} & \ja{ユニックス} & \ja{UIX} \\
Robota & T5\textsubscript{Megagon} & \ja{株式会社クリーン・アシスト} & \ja{株式会社クリーン・アジスト} & \ja{クリーン・アジスト}\\

\bottomrule
\end{tabular}
}
\end{table*}

\begin{table}[tb]
\centering
\caption{
Post-OCR Correction accuracy.
}
\label{tab:postocr_correction_accuracy}
\resizebox{0.8\columnwidth}{!}{%
\small
\begin{tabular}{c|r|r |r}
\toprule
Method      & Vision set  & Robota set & Average \\ \hline
Rule-based & 76.2 & 98.1  & 87.2\\ 
T5\textsubscript{Megagon} & 12.7 & 5.9 & 9.3\\
T5\textsubscript{Retrieva}  & 90.1  & 99.5& 94.8\\
\midrule
w/o correction & 72.0 & 98.7& 85.4\\
\bottomrule
\end{tabular}
}
\end{table}%

\subsection{Quantitative Evaluation}
Table~\ref{tab:postocr_correction_accuracy} shows the error correction accuracies by language model T5 and rule base. 
Based on the average scores, T5\textsubscript{Retrieva} achieves 94.8\%, which is an improvement of 9.4 points from the accuracy of 85.4\% before the correction. 
Next, rule-based error correction and the T5 \textsubscript{Megagon} achieved accuracies of 87.2 and 9.3 percent, respectively. 
It is confirmed that error correction accuracy depends on the pre-trained weights even when the same T5 and hyperparameters are used. 
This may be considered a result of differences in compatibility between the post-OCR correction task and the pre-trained weights and the conditions under which the pre-trained models were trained.
The experiments confirm that error correction by the well-selected pre-trained language model is effective on the average score, and that the rule base is also practical. 

The Vision set shows a similar trend to the average results, while the Robota set shows a different trend. 
Error correction by T5\textsubscript{Retrieva} improved the accuracy over the pure OCR results, while the other methods worsened the accuracy. 
When OCR accuracy is high, error correction may worsen it more than it improves it, so it is necessary to decide whether or not to correct the errors when applying the correction in practice.

\subsection{Qualitative Evaluation}
Table~\ref{tab:correction_result} shows before and after correction and GT examples of T5 models in each set. 
First, we can confirm that the T5 \textsubscript{Retrieva} in the Vision set can complete the missing "\ja{株式会社}," which means company type.
Also, it can be confirmed that while it can estimate a significant change in appearance from "Z" to "\ja{ほっかほっか亭}," it occasionally performs a different correction from GT. 
As for T5\textsubscript{Megagon}, we confirm that it tends to delete the first part of the input text, which is the reason for the incorrect result. 
We confirmed that the output trends differed significantly depending on the pre-trained weights, even for models with similar conditions.

The Robota set basically consists of correct OCR results, and it was confirmed that T5textsubscript{Retrieva} could produce identical output after correction without incorrect correction in the correct OCR recognition results.
On the other hand, T5\textsubscript{Megagon} tends to delete the beginning of the input text as in the Vision set, resulting in incorrect answers.

\section{Conclusion}
In this paper, we created a post-OCR error correction benchmark, JaPOC, to improve the OCR accuracy of Japanese vouchers. 
Based on the benchmarks, we constructed a simple baseline using the T5 language model and verified its effectiveness using two different pre-trained weights. 
Furthermore, we tested the effectiveness of the language models compared to the rule-based approach.
 The language model approach may be compatible with the task content and the pre-trained weights. 
The experiment showed that suited pre-trained models can effectively perform OCR error correction despite their simple structure.

 \bibliographystyle{splncs03_unsrt}
 \bibliography{article_new}
\end{document}